\documentclass[conference]{IEEEtran}

\makeatletter
\def\ps@IEEEtitlepagestyle{%
\def\@oddfoot{\mycopyrightnotice}%
\def\@evenfoot{}%
}
\def\mycopyrightnotice{%
{\footnotesize 978-1-6654-7095-7/22/\$31.00~\copyright~2022 IEEE\hfill}
\gdef\mycopyrightnotice{}
}

\usepackage{blindtext}
\usepackage{eso-pic}
\IEEEoverridecommandlockouts
\usepackage{cite}
\usepackage{amsmath,amssymb,amsfonts}
\usepackage{url}
\usepackage{booktabs}
\usepackage{algorithmic}
\usepackage{graphicx}
\usepackage{textcomp}
\usepackage{xcolor}
\def\BibTeX{{\rm B\kern-.05em{\sc i\kern-.025em b}\kern-.08em
    T\kern-.1667em\lower.7ex\hbox{E}\kern-.125emX}}
    
\usepackage{eso-pic}
\newcommand\AtPageUpperMyright[1]{\AtPageUpperLeft{%
 \put(\LenToUnit{0.17\paperwidth},\LenToUnit{-2cm}){%
     \parbox{0.9\textwidth}{\raggedleft\fontsize{8}{11}\selectfont #1}}%
 }}%
\newcommand{\conf}[1]{%
\AddToShipoutPictureBG*{%
\AtPageUpperMyright{#1}
}
}

\begin{document}
\title{\vspace*{1cm} Experiments on Anomaly Detection in Autonomous Driving by Forward-Backward Style Transfers

\thanks{This work results from the KIGLIS project supported by the German Federal Ministry of Education and Research (BMBF), grant number 16KIS1231.}

}

\author{\IEEEauthorblockN{Daniel Bogdoll\IEEEauthorrefmark{2}\IEEEauthorrefmark{3}\textsuperscript{\textasteriskcentered},
Meng Zhang\IEEEauthorrefmark{2}\textsuperscript{\textasteriskcentered},
Maximilian Nitsche\IEEEauthorrefmark{3},
J. Marius Zöllner\IEEEauthorrefmark{2}\IEEEauthorrefmark{3}
}

\IEEEauthorblockA{\IEEEauthorrefmark{2}FZI Research Center for Information Technology, Germany.
Email: bogdoll@fzi.de}
\IEEEauthorblockA{\IEEEauthorrefmark{3}Karlsruhe Institute of Technology, Germany.}
}

\maketitle

\begingroup\renewcommand\thefootnote{\textasteriskcentered}
\footnotetext{These authors contributed equally}
\endgroup

\conf{\textit{  Proc. of the International Conference on Electrical, Computer, Communications and Mechatronics Engineering  (ICECCME) \\ 
16-18 November 2022, Maldives}}

\begin{abstract}
Great progress has been achieved in the community of autonomous driving in the past few years. As a safety-critical problem, however, anomaly detection is a huge hurdle towards a large-scale deployment of autonomous vehicles in the real world. While many approaches, such as uncertainty estimation or segmentation-based image resynthesis, are extremely promising, there is more to be explored. Especially inspired by works on anomaly detection based on image resynthesis, we propose a novel approach for anomaly detection through style transfer. We leverage generative models to map an image from its original style domain of road traffic to an arbitrary one and back to generate pixelwise anomaly scores. However, our experiments have proven our hypothesis wrong, and we were unable to produce significant results. Nevertheless, we want to share our findings, so that others can learn from our experiments.
\end{abstract}


\section{Introduction}
\label{sec:intro}

Driven by the technology of deep neural networks for computer vision, tremendous advances have been seen in the field of autonomous driving. For instance, various deep learning models~\cite{https://doi.org/10.48550/arxiv.1905.05055} are nowadays able to provide outstanding performances for the task of object detection. However, in order to achieve effective deployment for real-world scenarios, reliability issues must be overcome. That is to say, under rare or unknown conditions, an autonomous vehicle is required not only to be able to identify the object classes from the training dataset, but also to detect atypical objects that have not been included in the training set. Anomaly detection, therefore, is an active topic in the research field of autonomous driving. As shown in~\cite{bogdollAnomalyDetectionAutonomous2022}, there are five main strategies to detect anomalies: Confidence scores, reconstructions, generative approaches, feature extraction methods and prediction errors. The approach presented here falls into the category of reconstructive methods, in which an attempt is made to produce a reconstruction of an original image via a certain process, with which, differences from the original image can be classified as anomalies.

\begin{figure}[t]
    \centering
    \includegraphics[width=\linewidth]{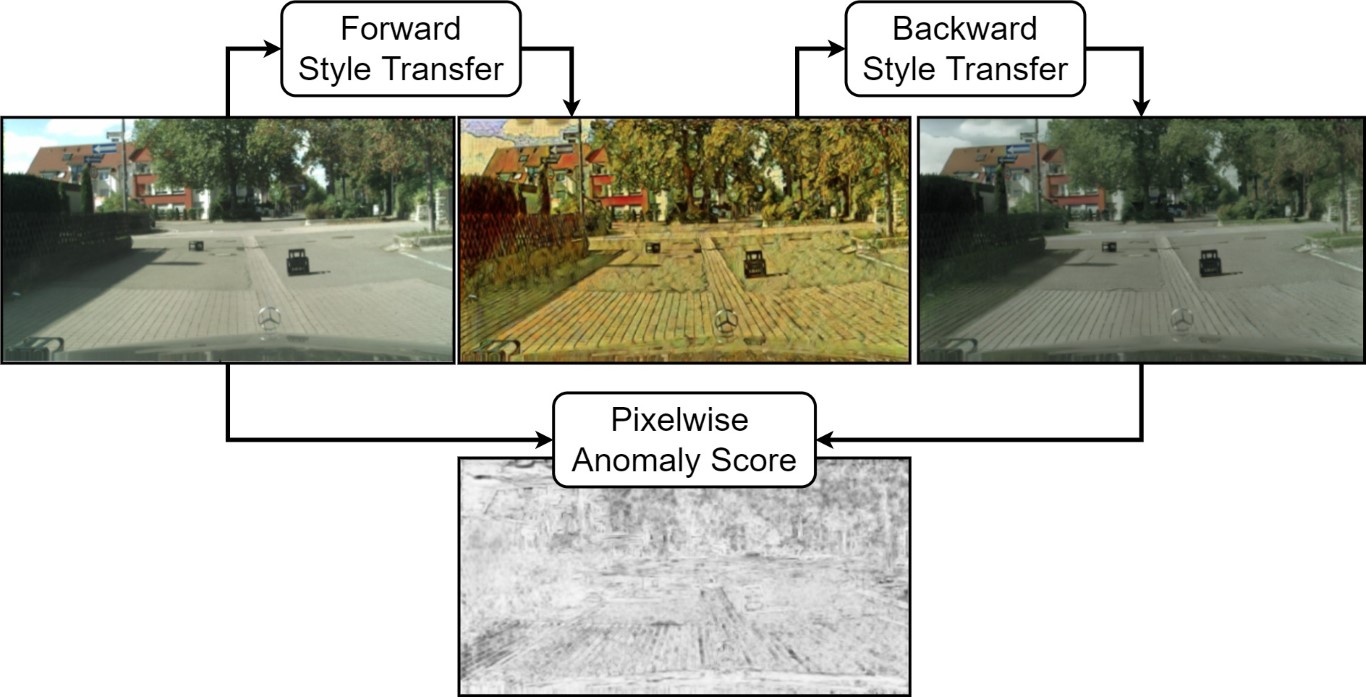}
    \caption{Overview of the novel anomaly detection approach. An image with anomalies is transferred into a stylized version of it with a forward style transfer, based on a target domain of choice. The resulting image is then transformed back to its original domain. Finally, a difference map between the original and the re-created image is computed, which can be interpreted as a map of pixel-wise anomaly scores.}
    \label{fig:framework}                          
\end{figure}

Inspired by approaches in this category, we came up with the idea to use generative models for style transfer based reconstructions, as shown in  Figure~\ref{fig:framework}. We have formulated our hypothesis as follows: Transferring an image from the road traffic domain, which includes anomalies, such as unknown objects, to another domain and back to the original domain, will lead to a resulting image that is similar to the original in the domain-typical aspects and deviates from it in the anomalous regions, which allows for the detection of anomalies. This hypothesis was made under the assumption that anomalies might disappear or get distorted through two successive style changes, which would allow for the detection of such. This hypothesis to detect anomalies is based on an assumption with respect to reconstruction errors with generative models, where “…normal samples are located on a manifold and all anomalous samples are located outside.”~\cite{anomalyscore_2019}.

For the experiments, we used the KITTI~\cite{Geiger2012CVPR} and Cityscapes~\cite{cordts2016cityscapes} datasets, and qualitatively evaluated the detection of anomalies with the Fishyscapes~\cite{blum2019fishyscapes} and Lost and Found~\cite{pinggera2016lost} datasets. However, our experiments have proven our hypothesis wrong and have shown that the concept as we have implemented it does not seem suitable for anomaly detection. To share this finding with the scientific community, we present our hypothesis, experiments, and negative results, as is often suggested~\cite{fanelli2012negative,nr_editage,nr_elsevier,nr_nature2,nr_nature,nr_ieee}, but rarely practiced.

\section{Related Work}
\label{sec:sota}

Most of the current research in anomaly detection for autonomous driving focuses on camera-based approaches~\cite{bogdollAnomalyDetectionAutonomous2022}, which we will present in more detail in the following section. Subsequently, we will give an overview of style transfer methods.

\subsection{Camera-based Anomaly Detection}
 Modern anomaly detection methods that work on camera data can be split into generative, reconstructive, and confidence-based approaches~\cite{bogdollAnomalyDetectionAutonomous2022}. The latter detect anomalies by thresholding a derived uncertainty measure of the model~\cite{kendallBayesianSegNetModel2016,CornerCaseDetection2021}. Moreover, Du~et~al.~\cite{duVOSLearningWhat2022} focus on the calibration of the neural network's confidence score to differentiate in- and outliers more thoroughly. For this purpose, they shape the decision boundary via virtual outliers sampled from a learned feature distribution~\cite{duVOSLearningWhat2022}. Reconstructive approaches on the other hand try to resynthesize the normality of a driving scene. Here, normality can be understood as what a model learned, given certain training data. Anomalous objects or events are then detected by comparing the real and reconstructed scene, resulting in a reconstruction error. However, modern generative adversarial networks (GANs) have shown outstanding reconstruction performance even on unseen data \cite{Bogdoll_Compressing_2021_NeurIPS}. In order to predict normality in autonomous driving, various machine learning algorithms are being used~\cite{bogdollAnomalyDetectionAutonomous2022}, like restricted Boltzmann machines~\cite{creusotRealtimeSmallObstacle2015}, autoencoder architectures~\cite{ohgushiRoadObstacleDetection2021}\cite{vojirRoadAnomalyDetection2021a}, or normalizing flows~\cite{blumFishyscapesBenchmarkMeasuring2021}. The NFlowJS model proposed by Grcić~et~al.~\cite{grcicDenseAnomalyDetection2021}, however, uses normalizing flow to synthesize outliers during training and to sensitize the model for potential anomalies. Other generative approaches use GANs to resynthesize the input image~\cite{haldimannThisNotWhat2019}\cite{nitschOutofDistributionDetectionAutomotive2021} and detect anomalous objects on the road, e.g., by the perceptual loss~\cite{lis2019detecting}. While there also exist methods that are based on feature extraction~\cite{bogdollAnomalyDetectionAutonomous2022}, these are typically older, which is why we excluded them from this overview.

\subsection{Image-to-Image Style Transfer}
Benefiting from the powerful feature extraction abilities of deep neural networks, image style transfer has been an active research area in image translation. Specifically, the aim of style transfer is to render an input image into a new image, which has a similar style and texture as a style image from a target domain. As pioneers of style transforms, Gatys~et~al.~\cite{gatys2016image} firstly proposed their method Neural Style Transfer (NST) to synthesize an image to match the style of a painting by an artist. Notably, separate control of the content and style of the picture is achieved by introducing content and style loss functions. However, due to its application to single image pairs, this approach cannot be used for large-scale style conversions. Motivated by Neural Style Transfer, another style transfer model, introduced by Johnson~et~al.~\cite{johnson2016perceptual}, performs much more efficiently by optimizing an image transform net.

In addition to ordinary neural network models, the emergence of GANs~ \cite{goodfellow2014generative,goodfellow2016nips} opens up more possibilities in the area of style transfer. As the name suggests, the core idea of GANs is to create desired images by simultaneously training two opposing neural networks. While a generator is responsible for synthesizing more realistic images under the guidance of a discriminator, the discriminator is trained to distinguish fake images from real images more accurately. Specifically, there are two well-known generative models for the task of image translation. With the help of adversarial and perceptual losses, the framework Pix2Pix~\cite{isola2017image} learns the mapping from a source domain to the target domain by training a model with paired training data. Differently, CycleGAN~\cite{CycleGAN2017} employs a cycle consistency loss~\cite{5596017}, as also applied by DualGAN~\cite{yi2017dualgan}, to conduct image-to-image translation under the unsupervised setting, making style transfers more flexible. Similar works include DiscoGAN~\cite{discogan}, GANimorph~\cite{Gokaslan2018}, ACL-GAN~\cite{zhao2020aclgan}, and AttentionGAN~\cite{tang2021attentiongan}. Besides, Park~et~al.~\cite{park2020contrastive} firstly applied contrastive learning~\cite{https://doi.org/10.48550/arxiv.2005.04966} for unpaired image-to-image translation. Interestingly, the model is trained to learn the invariance between the patch of the generated image and its corresponding patch in the input image by using the InfoNCE loss~\cite{van2018representation}.

\section{Approach}
\label{sec:method}
Our hypothesis to detect anomalies is based on an assumption in respect to reconstruction errors with generative models, where “…normal samples are located on a manifold and all anomalous samples are located outside. Since the manifold can be learned only where the training data lie…”~\cite{anomalyscore_2019}, we were curious if generative models for style transfers might perform worse in areas of data points outside of their learned distribution. Thus, the core idea of our approach is to perform anomaly detection by leveraging the impact of successive style transformations, as visualized in Figure \ref{fig:framework}. The forward style transfer is responsible for the conversion of an input image into a stylized version of it. The backward style transfer is then tasked to transform the image back to its original style domain. Finally, a pixel level comparison between the input and re-constructed image is conducted to recognize anomalies in the input image. 

\textbf{Forward Style Transfer}
By forward style transfer, we refer to the process of transferring an input image to an arbitrary style domain. Thereby, the input image will be rendered in the style of the target image, while preserving its content.

\textbf{Backward Style Transfer}
The aim of backward style transfer is to transform the stylized image, which is the output from the forward style transfer, back to the original style of the input image. 

\textbf{Pixelwise Anomaly Detection}
Finally, a pixel-by-pixel comparison is made between the input image and the newly synthesized image. Based on our hypothesis, we expected to detect anomalies in the input image through comparison. While we are aware that such a direct comparison between the re-synthesized image and the original is not sufficient for anomaly detection due to noise~\cite{di2021pixel}, spikes in such difference maps are a sufficient indicator for the general suitability of the approach.

\section{Evaluation}
\label{sec:eval}
In this section, we present qualitative results of our approach. We performed three different sets of experiments and show the resulting pixelwise anomaly scores for multiple input images. As we performed the experiments as a first step to determine whether the methodology is suitable for anomaly detection, we omitted a quantitative analysis afterwards.

\subsection{Data Selection}
Since the training data should reflect normality in road traffic, we first chose to utilize the well established KITTI~\cite{Geiger2012CVPR} dataset and also performed experiments with the much larger Cityscapes dataset~\cite{cordts2016cityscapes}. For the following evaluation, datasets with anomalies were necessary. While there exist many datasets in the field of autonomous driving~\cite{Bogdoll_Addatasets_2022_VEHITS}, only a few of them are suited for the task of anomaly detection. We utilize data samples from the Fishyscapes~\cite{blum2019fishyscapes} and Lost and Found~\cite{pinggera2016lost} datasets to include both real and augmented anomalies. As the Lost and Found dataset is part of the Fishyscapes suite, we might refer to both as Fishyscapes. Both provide ground-truth labels for out-of-domain pixels, which we consider anomalies. In autonomous driving, these are also often called corner cases~\cite{Bogdoll_Description_2021_ICCV} in the category unknown object~\cite{9304789}. These unknown objects can typically also be seen clearly in the images directly. Figure~\ref{fig:model_evaluation} shows a selection of animals and small objects. For the art styles used for the forward passes, we selected a wide variety of styles, including art from Van Gogh, Giles and pieces in the Ukiyo-e style.

\subsection{Model Selection}
We have performed experiments with two different image style transfer methods. First, we used CycleGAN~\cite{CycleGAN2017}, which is an established framework for unpaired image translations. For model-based style transfers, it is important to choose a model that can be trained with unpaired images as training data, contrary to approaches such as Pix2Pix~\cite{isola2017image}, where paired images are necessary. Among those, CycleGAN~\cite{CycleGAN2017} is the most prominent representative. While there are newer models, CycleGAN is optimal for experiments, since high quality open-source code~\cite{cyclegan_code} is available and many works have used it already, making it a mature method.\\

Second, we utilized the Neural Style Transfer approach~\cite{gatys2016image}, which is regarded as one of the early milestones in the field of style transformations. While CycleGAN is able to generate stylized images with a high quality, we have no dynamic control over how much an image should be transformed. This becomes possible with NST, so that we were able to examine the influence of conversion degrees. Based on convolutional neural networks (CNN), the Neural Style Transfer Algorithm is able to render a random input image in the style of a target image by leveraging the VGG19 model~\cite{simonyan2014very}, which has been widely used for feature extraction in various tasks. Moreover, the highlight of the algorithm is controlling the transfer degrees through two weight parameters of the loss functions: content weight and style weight. On the one hand, a content image would be always compared with the generated image, in order to make sure the semantic content is retained in the process of stylization. Specifically, the content structure can be preserved through minimizing the mean squared error of feature maps between a content image and the input image. On the other hand, through computation of the Gram matrix, a style image in the target domain would be always compared with the generated image, such that the synthesized image will have a similar style as the given style image. Thus, it is a tradeoff problem for the task of combining the content image with another style image, where it is impossible to meet the two constraints perfectly at the same time. For instance, the output image could have the texture of the style image, but lose some content when a larger style weight is set. On the contrary, a much larger content weight would result in that the synthesized image is further away from the target style. 

\subsection{Style Transfer with CycleGAN}
\label{sec:cyclegan}
In this first experiment, we have utilized CycleGan for both the forward and backward style transfers. First, we will give an overview of the details of the forward style transfer.

\subsubsection{Forward Style Transfer}
For this experiment, instead of training a CycleGAN model from scratch, a pre-trained model was directly used for the forward style transfer. Although the pre-trained CycleGAN model~\cite{cyclegan_code} was trained on paintings by Van Gogh as the target domain and ordinary photos as the source domain, it has shown to have good generalization capabilities on various datasets, such as KITTI, Cityscapes, as well as Fishyscapes and Lost and Found.

\subsubsection{Backward Style Transfer}
For the backward style transfer direction, however, the observed performance of the pre-trained GAN was not sufficient. Therefore, we chose to train the model from scratch. Notably, in order to perform the task of style transfer with a CycleGAN model successfully, images from both the source and target domain are required to be fed into the model for training. Specifically, paintings by famous artists are commonly used as source images and ordinary photos are taken for the target domain. For achieving better performance of the backward transfer, we trained our model on several data sources, which are presented in Table~\ref{table:training}.
    
\begin{table}[h]
\centering
\resizebox{\columnwidth}{!}{%
\begin{tabular}{@{}lllll@{}}
\toprule
Group & Source Domain                   & Target Domain     \\ \midrule
A     & Paintings by Van Gogh           & KITTI             \\
B     & Paintings by Van Gogh           & KITTI + photos    \\
C     & Ukiyo-e style paintings         & KITTI             \\
D     & Ukiyo-e style paintings         & KITTI + photos    \\
E     & Van Gogh stylized KITTI         & KITTI             \\
F     & Van Gogh stylized Cityscapes    & Cityscapes        
\end{tabular}%
}
\caption{Data sources used for the training of CycleGAN models}
\label{table:training}
\end{table}

We firstly conducted four sets of trainings, shown as groups A to group D. Paintings by Van Gogh were adopted as the source domain images in both groups A and B. In group A, we used images from the KITTI dataset as the target domain, whereas a combination of KITTI data and ordinary photos served as target style images in group B. Additionally, we employed the Japanese painting style Ukiyo-e~\cite{ukiyoe} as the source style in group C and D. For the target domain, group C used KITTI data as done for group A, and group D combined KITTI with ordinary photos as in group D. However, this set of experiments did not perform well, even though we observed slightly better results when using Van Gogh style images and KITTI data. Here, two problems occurred. On the one hand, some noise was constantly present, which we attribute to the special texture of Van Gogh paintings. On the other hand, the performance of the model was erratic, resulting in the occasional lack of much difference between transformations and the corresponding input images. After this analysis, we found two directions to resolve the problem. Firstly, rather than directly using paintings from artists, we took stylized images, which were the output from the forward style transfer network, to train our model with more domain-specific data. Secondly, there were too few images from the source domain for training. With the newly selected data source, we were able to increase the number of stylized images significantly and performed another two trainings, as shown in groups E and F. In group E, Van Gogh stylized KITTI samples were used for the source domain, and KITTI samples were used for the target domain. In group F, we followed the same approach, but used Cityscapes data. 

\subsubsection{Pixelwise Anomaly Detection}
\begin{figure}[h]
    \centering
    \includegraphics[width=\linewidth]{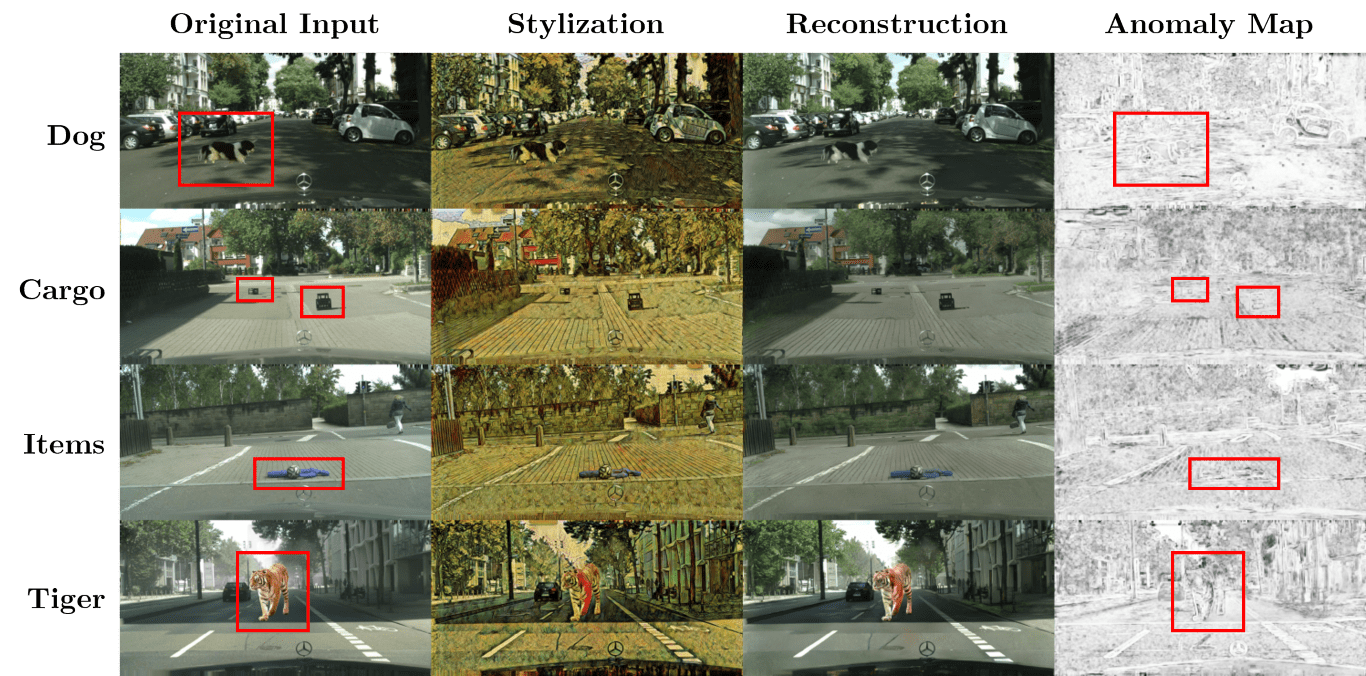} 
    \caption{Evaluation results of the model trained with Van Gogh stylized Cityscapes images as the source domain and Cityscapes samples as the target domain. Four different types of anomalies are included in the Fishyscapes scenes, which are highlighted in red.}
    \label{fig:model_evaluation}
\end{figure}

For the evaluation, an input image chosen from the Fishyscapes dataset was used as the input. With the final approach as shown in group F of Table~\ref{table:training}, we were able to generate images with the backward style transfer pass that were consistently close to the original ones, as shown in Figure~\ref{fig:model_evaluation}. 

This observed performance improvement on the Fishyscapes dataset is largely due to the fact, that Fishyscapes is an extension of the original Cityscapes datasets, so the general domain remains unchanged. After the conversion of the input into a stylized version of it, based on the forward style transfer, we reconstructed the input image based on the backward style transfer. The re-synthesized images are generally very close to the original ones, which was a wanted effect. However, as visible in the last column, we did not observe any deformations in the area of the anomaly. 

\subsection{Style Transfer with NST}
\label{sec:nst}
For the second experiment, we used the Neural Style Transfer method for both the forward and backward style transformations. In the CycleGAN framework, weight parameters are updated iteratively, until the model converges. Contrary, with NST, it is a single image that is optimized in the training process. Thus, only two images are necessary for the process: a content image and a style image. Therefore, we directly utilize Fishyscapes data samples for the style transfers.

\subsubsection{Forward Style Transfer}
 For the forward style transfer, a sample from Fishyscapes is used as the content image, and a painting from Van Gogh is taken as the style image in the optimization process. Content weight and style weight are two important parameters, that can be set to control how much the image will be stylized. As recommended by reference implementations, 
 ~\cite{nst_code_1}\cite{nst_code_2}, the content weight is kept unchanged at $1e5$, such that the process of style transfer can be easily controlled by only changing the style weight.

 \begin{figure}[h]
    \centering
    \includegraphics[width=\linewidth]{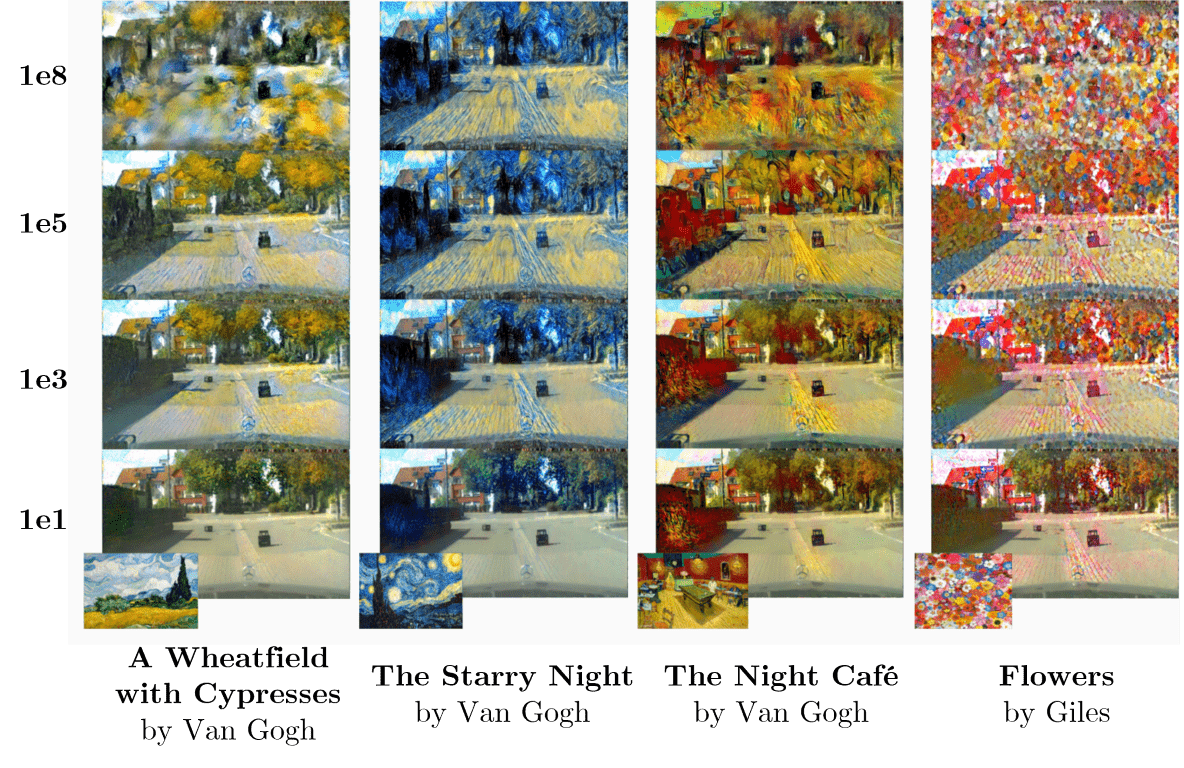} 
    \caption{Style transfers with NST with a scene from the Fishyscapes dataset with lost cargo as anomalies. Each column represents a different style and each row a different style weight.}
    \label{fig:5 styles}                
\end{figure}
 
To determine, which degree of style transfer is most likely to lead to distortions of the content, especially anomalies, during the forward style transfer, we performed a set of small experiments with the style weight. As visualized in Figure~\ref{fig:5 styles}, we performed experiments with four different styles~\cite{art1,art2,art3,art4}, and four different style weights. The larger the style weight, the better the generated image matches the style image. Since a style weight of $1e8$ often led to a nearly complete loss of the content information, we utilized a value of $1e5$ for the further experiment to maximize the effect of the style transfer, while retaining the content.

\subsubsection{Backward Style Transfer}
To convert the stylized Fishyscapes images back to their original style of a regular street scenario, we chose an image from the Cityscapes dataset as the style image. Importantly, this image from Cityscapes does not contain any anomalies, such that it only works as a style guidance, avoiding an unexpected influence on the content part.

\subsubsection{Pixelwise Anomaly Detection}
\begin{figure}[h]
    \centering
    \includegraphics[width=\linewidth]{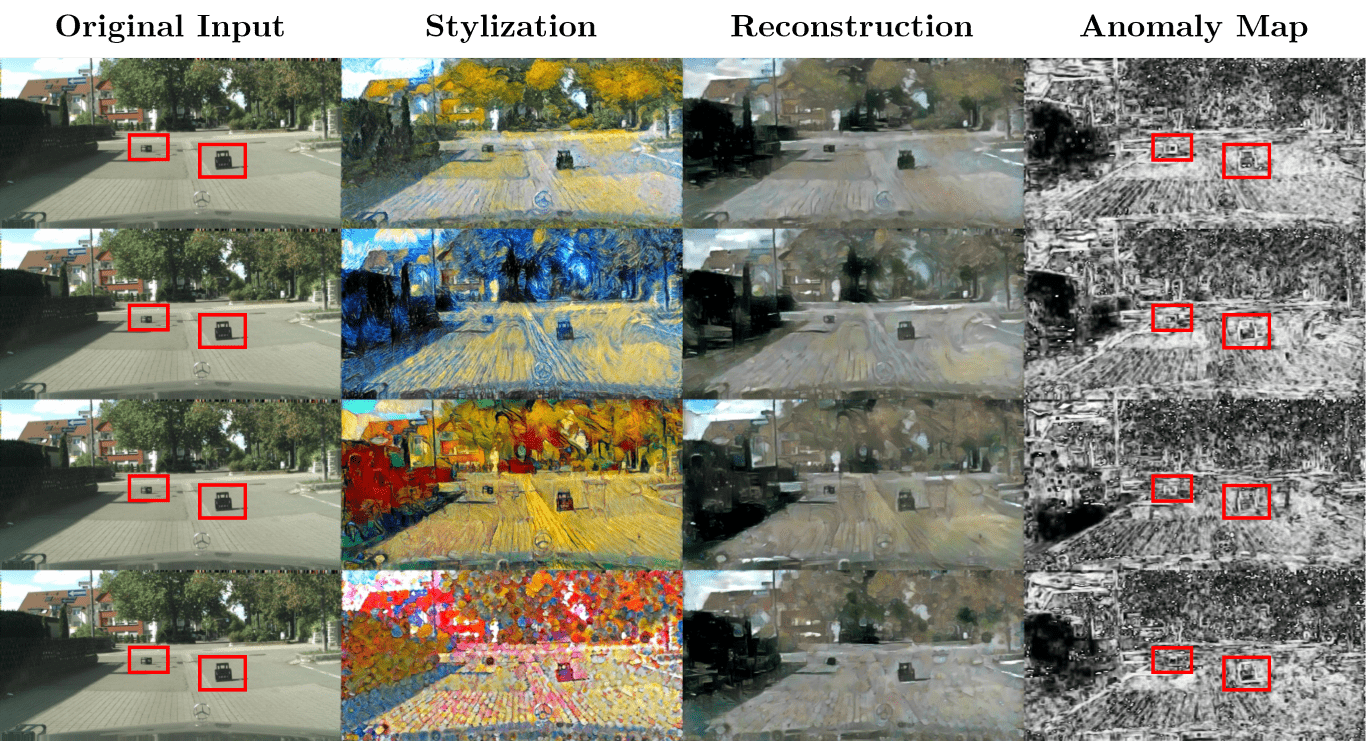} 
    \caption{Evaluation of Neural Style Transfer with a content weight and a style weight of $1e5$ each. A sample scene from the Fishyscapes dataset is used as the input. Each row shows the process with a different style. Anomalies are highlighted in red.}
    \label{fig:NST_combination_results}
\end{figure}

Illustrative results of the second experiment are shown in Figure~\ref{fig:NST_combination_results}. Here, we utilized the same styles as introduced in Figure~\ref{fig:5 styles}. To highlight the effects of the different styles on the reconstruction and anomaly detections, we show only one scene from the Fishyscapes dataset.

 Here, it becomes visible that the results of the backwards style transfers differ less than the differences of the stylized images might suggest. Again, the anomalies remain well visible in the reconstruction. The anomaly map in the last column shows no spikes in these regions, but shows a high level of difference values in general.

\subsection{Hybrid Style Transfer}
\begin{figure}[h]
    \centering
    \includegraphics[width=\linewidth]{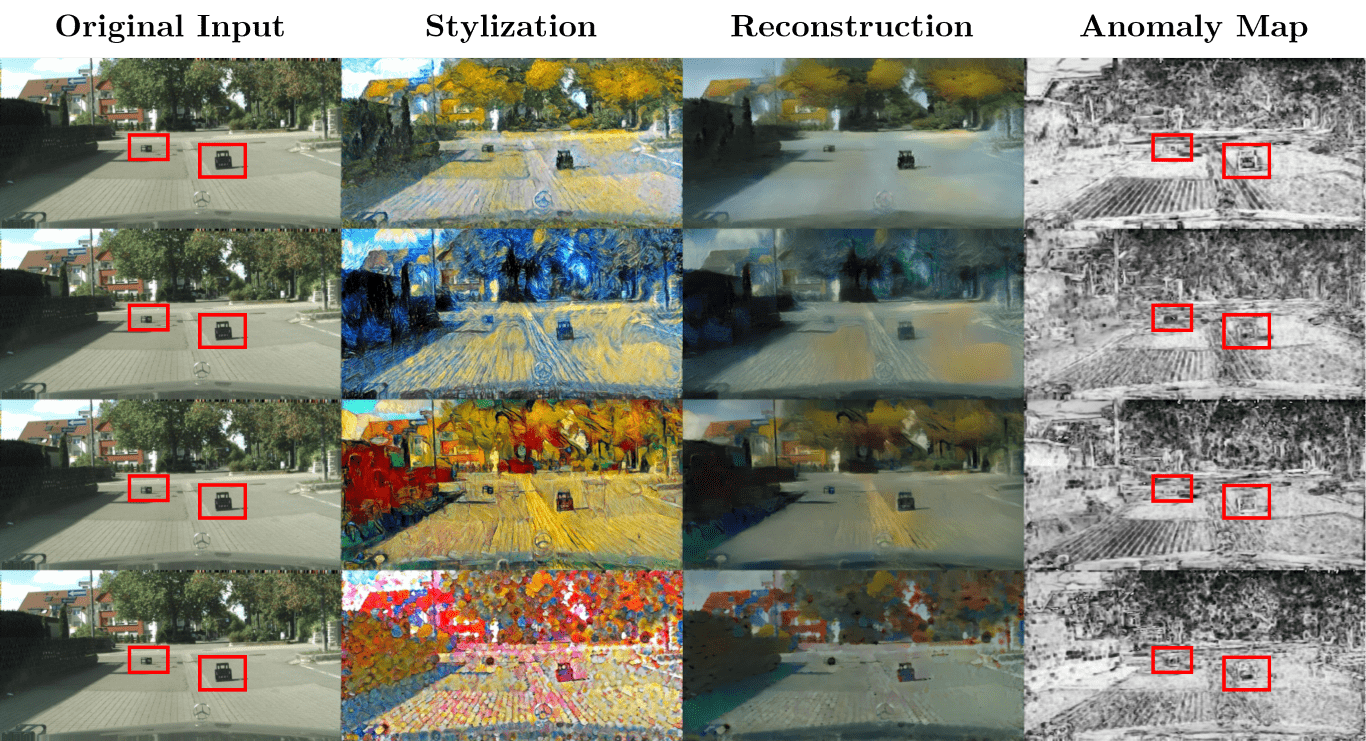} 
    \caption{Evaluation of the hybrid approach with a sample scene from the Fishyscapes dataset. Here, NST is responsible for the forward style transfer. The backward style transfer was carried out by our CycleGAN model. Each row represents a different style. Anomalies are highlighted in red.}
    \label{fig:pretrained_combination_results}
\end{figure}

For the third and final experiment, we combined both approaches. Here, we performed the forward style transfer with the Neural Style Transfer method, while we used our CycleGAN model, trained with Cityscapes data, for the backward style transfer. For this experiment, we followed the intuition that the NST allows for a better stylization control for the forward pass, while the CycleGAN enables higher quality results for the backward pass. As we have explained both passes in the previous Sections~\ref{sec:cyclegan} and \ref{sec:nst}, here we describe only the results, which are visualized in Figure~\ref{fig:pretrained_combination_results}. It can be seen, that with such a variety of styles, the GAN, which was only trained with one style, was not capable to perform the backward style transfer properly. Here, a leakage of the applied style to the reconstruction occurs with styles, which were not used for the training of the GAN. Regarding the anomaly detections, we observe a similar behavior as in the previous experiments. The amount of noise is, compared to the other two experiments, in a medium range.

With this presented set of three comprehensive experiments, we examined the influence of training data, style data, anomaly data, style weights, and style transfer models for the purpose of anomaly detection based on successive style transfers. While the results showed high-quality reconstructions, the style transfers unfortunately reconstructed the areas with anomalies similarly to the rest of the images. Thus, the difference maps showed primarily noise, which makes the detection of anomalies with this method unlikely. Specifically, our experiments have placed a focus on the comparison of  CycleGAN, Neural Style Transfer, and a hybrid combination of both for the Forward and Backward Style Transfers parts. The results differed in the value distribution of the difference maps and the amount of noise, but showed no sufficient differences for the pixels of the anomalies.
\section{Conclusion}
\label{sec:conclusion}
The aim of our work was to investigate whether our proposed approach of successive style transfers is suitable as a novel method for anomaly detection in the domain of autonomous driving. Therefore, we performed a set of three extensive experiments. Within these, we transferred images from the road traffic domain to multiple art domains with a set of forward style transfer methods and then transferred the results back to the original domain. We compared these reconstructions with the original images to calculate a pixelwise anomaly score. Our experiments revealed, that anomalies were still consistently visible in the reconstructed image, which did not lead to high anomaly scores in the difference maps within the areas of the anomalies. Based on our analysis visualizations, these negative results should serve as inspiration and knowledge to support following research in the field of anomaly detection in autonomous driving.

{\small
\bibliographystyle{IEEEtran}
\bibliography{ref}
}

\end{document}